\documentclass[conference]{IEEEtran}
\IEEEoverridecommandlockouts
\usepackage{cite}
\usepackage{amsmath,amssymb,amsfonts,amstext}
\usepackage[algoruled,linesnumbered,ruled,lined, noline, boxed]{algorithm2e}
\usepackage{xcolor}

\SetCommentSty{mycommfont}
\usepackage{graphicx}
\usepackage{textcomp}
\usepackage{xcolor}
\usepackage{caption}
\usepackage{subcaption}
\usepackage{tabularx}
\def\BibTeX{{\rm B\kern-.05em{\sc i\kern-.025em b}\kern-.08em
    T\kern-.1667em\lower.7ex\hbox{E}\kern-.125emX}}
\begin{document}
{
\title{LogAnMeta: Log Anomaly Detection Using Meta Learning }
}

\makeatletter
\newcommand{\newlineauthors}{%
  \end{@IEEEauthorhalign}\hfill\mbox{}\par
  \mbox{}\hfill\begin{@IEEEauthorhalign}
}
\makeatother

\author{\IEEEauthorblockN{Abhishek Sarkar}
 \IEEEauthorblockA{\textit{Senior Researcher} \\
 \textit{Ericsson Research}\\
 Bangalore, India \\
 abhishek.sarkar@ericsson.com}
 \and
 \IEEEauthorblockN{Tanmay Sen}
 \IEEEauthorblockA{\textit{Data Scientist} \\
 \textit{BA BCSS Digital Services}\\
 \textit{Ericsson}\\
 Kolkata, India \\
 tanmay.sen@ericsson.com}
 \and
 \IEEEauthorblockN{Srimanta Kundu}
 \IEEEauthorblockA{\textit{Solution Architect} \\
\textit{BA BCSS Digital Services}\\
 \textit{Ericsson}\\
 Kolkata, India \\
 srimanta.kundu@ericsson.com} 
 \newlineauthors
 \IEEEauthorblockN{Arijit Sarkar}
 \IEEEauthorblockA{\textit{Senior Solution Architect} \\
\textit{BA BCSS Digital Services}\\
 \textit{Ericsson}\\
 Bangalore, India \\
 arijit.sarkar@ericsson.com}
 \and
\IEEEauthorblockN{Abdul Wazed}
 \IEEEauthorblockA{\textit{Solution Architect} \\
 \textit{BA BCSS Digital Services}\\
 \textit{Ericsson}\\
 Kolkata, India \\
 abdul.wazed@ericsson.com} 
 }
\maketitle
\begin{abstract} 
Modern telecom systems are monitored with performance and system logs from multiple application layers and components. Detecting anomalous events from these logs is key to identify security breaches, resource over-utilization, critical/fatal errors, etc.
Current supervised log anomaly detection frameworks tend to perform poorly on new types or signatures of anomalies with few or unseen samples in the training data. In this work, we propose a meta-learning based log anomaly detection framework (LogAnMeta) for detecting  anomalies from sequence of log events with few  samples. 
LoganMeta train a hybrid few-shot classifier in an episodic manner. The experimental results demonstrate the efficacy of our proposed method
\end{abstract}

\section{Introduction} \label{sec1}
Log analysis is widely utilized to reduce manual intervention for troubleshooting and maintenance of the software systems. Log analysis can be categorize into two parts one is log parsing and another is abnormality detection. For complexity and high volume of logs made traditional way of log parsing obsolete, which is heavily depends on regular expressions based log keyword extraction.
Zhu et. al \cite{zhu2019tools} reproduce and evaluate a series of heuristic and clustering based automated log parsers on 16 different log data sets. In our work we have used automated log parsing algorithm, Drain \cite{he2017drain}, to parse the raw logs, as it is relatively better in terms of efficiency and accuracy, among all other available log parsers. Now come into discussion of anomaly detection of log data. In recent years, the problem of anomaly detection  of logs (quantitative and sequential) are studied widely using supervised, semi-supervised or unsupervised manner by machine learning or deep learning based approach.
He et. al. \cite{he2016experience} used equal number of supervised and unsupervised learning based algorithms to detect anomalous logs. He et. al. \cite{he2017towards} represented log sequences as log count vectors and applied supervised learning algorithms to detect anomalies. 
Xu et. al. \cite{xu2009detecting} first introduce PCA based unsupervised log anomaly detection method. Du et. al. \cite{du2017deeplog} and Zhang. et. al. \cite{zhang2016automated} proposed  sequential unsupervised anomaly detection method. They have used LSTM to predict the next log keys and an anomaly is declared if the fraction of log keys are significantly different compared with the actual log keys. Zhang et. al. \cite{zhang2019robust} proposed a robust log anomaly detection framework, which represents each log events as a fixed-dimension semantic vector and used  an attention-based Bi-LSTM  model to detect anomalies.
Meng et. al. \cite{meng2019loganomaly}, combined both sequential and quantitative pattern of log events for anomaly detection, which can significantly reduce false alarms. They convert each log keys to a vector by sentence embedding (template2vec) method, which is an effective way  to extract the semantic information in the sequence of log keys. 
The existing log based anomaly detection frame work is not very suitable for new types of anomalies with few anomalous samples. To the best of our knowledge this is the first attempt to solve log anomaly detection problem using meta learning. The key contribution of our work is outlined as follows: \
\begin{itemize}
\item We propose a hybrid meta learning based classifier Proto-Siamese net which is train on few shot samples in an episodic manner to detect anomalous events in sequence of logs. 

\item Experimental results indicate that the proposed approach outperforms the baseline approaches. 
\end{itemize}
\section{Meta Learning}
In recent years, deep learning has evolved rapidly with varieties of algorithms. But the main challenge with deep neural networks is that it's required to feed large amount of samples to train the model and it performs poorly on few data  as well as on imbalanced data problem. The concept of few shot meta learning can handle those obstacles as it can imitate human like learning to learn new concepts (tasks) with few training instances with fast and efficiently. We combined two well known meta learning approaches, siamese and proto net, and proposed a hybrid Proto-Siamese neural network based meta learning approach to detect anomalies in log. This approach is described in the next section. Now we will briefly describe about siamese neural network and prototypical network. 

\begin{itemize}
\item \textbf{Siamese Neural Network:} 
Siamese neural network \cite{koch2015siamese}, is one of the vastly used few-shots learning algorithm, which can learn from very few data points and subsequently solve small data or class imbalance data problem. It is composed of two identical networks with same weights parameters and  it learns by finding similarity between two embedded input vectors. 
\item \textbf{Prototypical Network:} Like siamese net, prototypical network \cite{snell2017prototypical} is also another efficient, simpler few shot learning algorithm, which learn a metric space for performing classification task. It creates a  prototypical representation of each class and a query point is classified based on the
distance between the class prototype and the query point.
It does episodic training, by generating support set and query set. Support set is created by picking up few samples randomly from each class and train the network. 
Similarly, it performs classification task on a query point (new data point) by randomly picking a sample from the data set. 
\end{itemize}

\section{Model Architecture }

\begin{figure}[!htp]
\includegraphics[scale=0.4,trim={0 6cm 10cm 0}, clip]{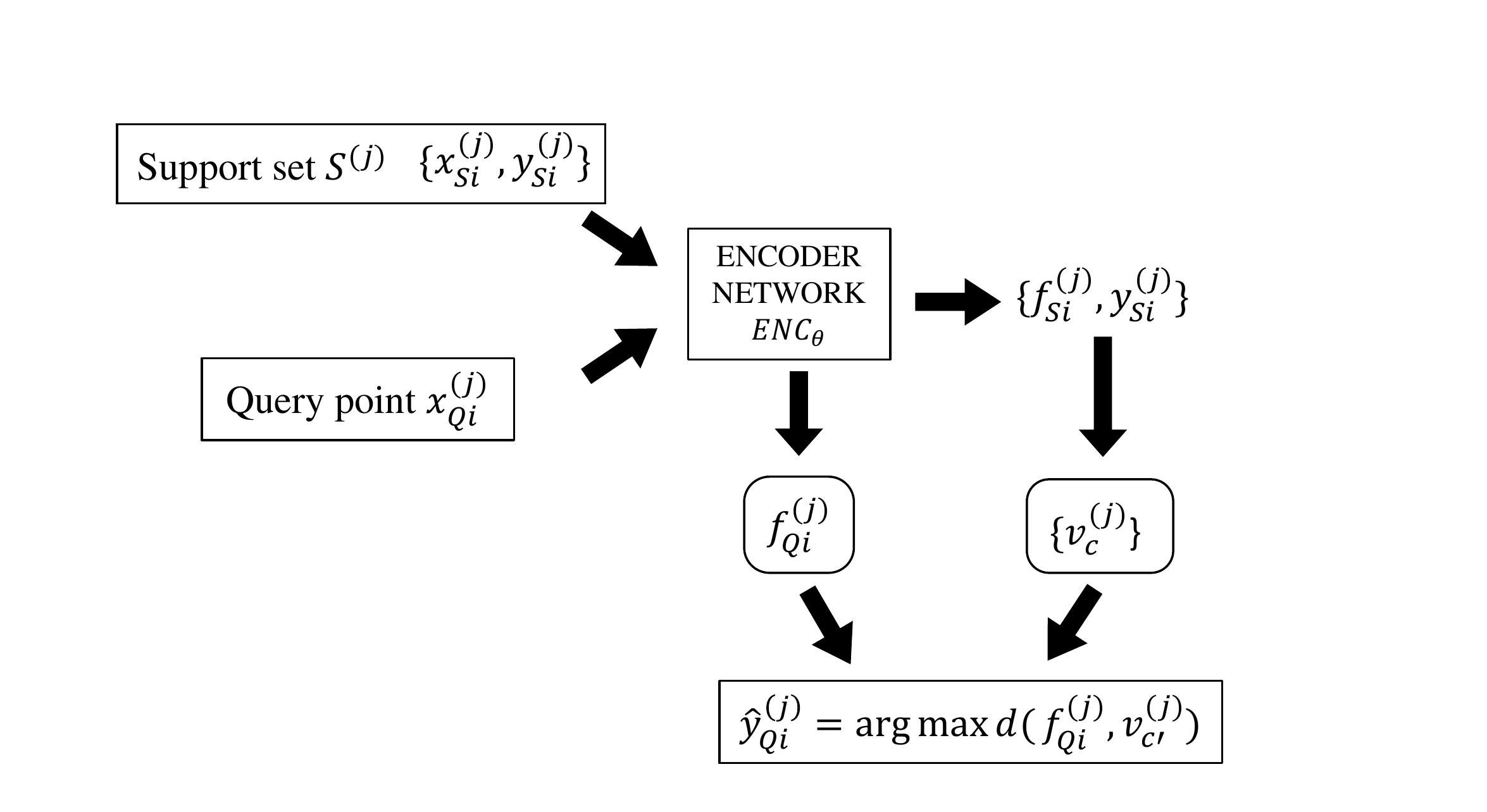}
\label{fig:arch}
\caption{The run-time architecture of the proposed hybrid proto-Siamese model. The model is trained by minimizing average of 2 loss function $L_{\rm{proto}}$ and $L_{\rm{triplet}}$  }
\end{figure}

Let $ D = \{x_i, y_i\}_{i=1}^{P}$ denote the overall training data where $x_i \in R^{d}$ are $d$-dimensional feature vectors and $y_i \in \mathcal{C} = \{0, 1, \ldots N_C \}$ are the corresponding labels. Examples with label $0$ are assumed to be normal data-points while non-zero labels are assumed to be different types of anomalies. To encourage the model to learn from a mix of normal and anomalous data points, we first partition the whole data set into normal data $Nor \subset D$, where $Nor = \{ \{x_i, y_i \} \in D | y_i = 0  \}$ and anomalous data $Ano \subset D$, where $Ano = \{ \{x_i, y_i \} \in D | y_i \neq 0  \}$. 
We set up the meta-learning problem as a $N$-way, $K$-shot problem, where the objective would be to output a $N$-way classifier from $K$-shot samples from each of the $N$-way classes. Each task $\tau_j$ in the meta-learning algorithm then is composed of a support set and a query set, $\tau_j = \{ S^{(j)}, Q^{(j)} \}$ where the support set $S^{(j)} = \{ x^{(j)}_{Si}, y^{(j)}_{Si} \}$ is constructed by mixing $K$-shot instances each of normal and anomalous samples such that they have $N$-way classes, such that $ y^{(j)} \in \mathcal{C}_0 \subset \mathcal{C}$ where $|\mathcal{C}_0| = N$  and $|S^{(j)}| = N \times K$. The query set $Q^{(j)} = \{ x^{(j)}_{Qi}, y^{(j)}_{Qi} \}$ is constructed in a similar manner. We define an encoder network $ENC_\theta$ where $\theta$ are the learnable weights of the network. Given an input sample $x_i$, $ENC_\theta$ produces an embedding $f_i = ENC_\theta(x_i) \in R^{d_{emb}}$, where $d_{emb}$ is the embedding dimension.  With every example in the support set embedded using $ENC_\theta$, we compute class-specific prototypes, $v^{(j)}_c = \frac{1}{|S^{(j)}_c|}\sum_{(x_i, y_i) \in S^{(j)}_c} f_i$, where $S^{(j)}_c = \{(x_i, y_i) \in S^{(j)} | y_i = c \in \mathcal{C}\}$.


To classify a new query point $x^{(j)}_{Qi}$, we compute the distance between the query point and the $\{v_c\}$ and assign the class of the closest prototype. Thus, $\hat{y}^{(j)}_{Qi} = \arg \max_{c' \in \mathcal{C}} -d(f(x^{(j)}_{Qi}), v_{c'})$, where $d$ is a suitably chosen distance function. 

\subsection{Loss Function}
\noindent 

Thus our model needs to learn good representations $\{ f_i\}$ of $\{x_i\}$. We train the parameters $\theta$ of the network $ENC_\theta$ using 2 loss functions. 
First, the loss function based on computing the nearest prototype is as follows: 

\begin{equation}
 L_{\rm{proto}} =  \frac{1}{|Q^{(j)}|} \sum_{Q^{(j)}} \sum_{c \in \mathcal{C}_0} I(y^{(j)}_{Qi} = c) \log \mathrm{Softmax}(-d(f(x^{(j)}_{Qi}), v_{c'}))
\end{equation}


Second, to encourage the model to better discriminate between features from normal sequence of logs and varying types of anomalous sequence of logs given only a few examples of each class. We employ a siamese style triplet loss, where the model explicitly acquires discriminative properties given triplets of anchor, positive and negative examples. This loss is defined as
\begin{equation}
    L_{\rm{triplet}}(x_a, x_p, x_n) = \frac{1}{N_{\rm{trip}}} \sum_{i} \max(d(x^a_i, x^p_i) - d(x^a_i, x^n_i) + \alpha, 0)
\end{equation}
where $\{x^a_i, x^p_i x^n_i \}_i$ are sequences of $N_{\rm{trip}}$ triplets chosen from the support set $S^{(j)}$, such that $y^a_i = y^p_i$ and $y^a_i \neq y^n_i$, $d$ is a suitably chosen distance function and $\alpha$ is a margin.

\subsection{Model Training}
Thus the combined loss function 
\begin{equation}
    L_{\rm{hybrid}} = \frac{1}{2} (L_{\rm{proto}} + L_{\rm{triplet}})    
\end{equation}
is used to train the entire model end to end. The training process starts off by creating different tasks, each task consisting of support and query sets (also called meta-train and meta-validation sets). Then for each task, the model produces embeddings of all the examples in the support set and the query set.  From the examples in the query set, it computes the prototypes embeddings for each class in the query set. Then, for each  example in the query set it computes the log softmax over the distances between the prototypes and the embedding of the example.  This operation allows us to compute the proto-type loss $L_{\rm{proto}}$. Within the same task, the training then selects a small number of triplets containing anchor, positive and negative examples at random to compute the triplet loss $L_{\rm{triplet}}$





\section{Experiments and Results}




\subsection{Data}
The data we use is a private dataset of logs collected within Ericsson's internal logging systems. This log dataset is then cleansed and we extract templates out of the data using Drain algorithm 
The data is then annotated by subject matter experts on whether 5-minute sequences of logs are anomalous or not. Once this annotated dataset is prepared, we convert the sequence of logs to a rectangular dataset where each row of data is a sequence of logs of duration 5 minutes and each feature is the count of log templates present in that 5 minutes of log messages. This raw count-based dataset is then post-processed using tf-idf to produce our feature sets for our training model. The binary labels 0 and 1 in the original annotated data determine whether a sequence of logs is normal or anomalous. With the anomolous records do a further round of processing to determined 6 different types of sub-anomalies from the data. Combined with the normal class, the final labelled data has 31,800 rows, 7 classes and 495 features.  Out of the 7 classes, class-0 denotes normal sequences and account for about 93\% of the data, while the rest 8\% are distributed among the 6 different types of anomalies.

\subsection{Implementation Details} 

\subsubsection{EncoderNetwork}
The encoder network is a three layer sequential network with input dimension as 495. The intermediate 2 layers have 128 hidden units each while the output dimesion is 32. Thus, each 495-d feature vector is embedded onto a 32-d space. Each layer except the last is followed by a ReLU activation unit. All layers have dropout with $p=0.5$ as default. Total training time was 1.5 hours. 

\subsubsection{Hyper-parameters}
We train our hybrid proto-net model for 500 epochs with the AdamW optimizer, starting out with a learning rate of $1e-3$, placing equal weights of 0.5 on both loss functions. We employ a multi-step scheduler with a decay of 0.1 with milestones at epochs 150 and 450.

\subsubsection{Environment}
The experiments were carried out on a desktop with Intel(R) Core-i3 3.6 GHz CPU with 16g RAM running  Windows-10, Conda version 4.12.0 with Python 3.9.13,  Pytorch 1.10.2.

\subsection{Baseline}
We build a traditional supervised binary classifier as well as a multi-class classifier for baselining our hybrid algorithm. The binary classifier is a deep feed-forward net with 3 layers with the intermediate layers of size 128 and 64 with Relu activation and dropout with 0.5 as default.  The multi-class classifier has the same intermediate architecture but has an output layer of size 6.  Both models were trained to 100 epochs with the Adam optimizer with a learning rate of 1e-6 and a batch-size of 32.

\subsection{Results} 

Table \ref{tab:cmprbsln} compares our final model results with the baseline binary and multi-class classification models.
From Table \ref{tab:cmprbsln} we see that our model achieves a final accuracy of about 98.6\% on the held-out set outperforming the multi-class classification which achieved a validation accuracy of about 95.3\%.  The model achieves a comparable accuracy with the baseline binary classifier. The multi-class classifier sufferes significantly due to the severe imbalance of different classes of anomalies present in the training data.  Due to our meta-learning set up and our design of 2 explicit cost functions to tackle the problem, the hybrid model is able to circumvent the problem of severe class imbalance. 

Figure \ref{fig:rslts} presents the overall training regime of our proposed model. We see that both the components of the loss, the proto-loss $L_{\rm{proto}}$ and $L_{\rm{triplet}}$ are minimized. The model is able to learn both disciminative representations of the encoded log sequences as well learns to perform nearest neighbor classification. In the initial stages, we observe that the validation accuracy and both the loss outperform the training counterparts. We think this may be due to the specific split of the anomaly groups into training and validation steps. The classes in the validation split are slightly easier to learn from than the corresponding classes included in the train split. However this difference disappears mid-way into the training. This early discrepancy between the performance can be the focus of future work. Also notable is the variance in the loss curve for different epochs. This  variation is due to the nature of the meta-learning task set up where we perform a 2-way 2-shot classification. This issue can also be addressed in future work where we set up explicit meta-learning specific regularizers to combat high variance. 
To visualize the quality of the embeddings produced by our model, we compress the 32-dimensional embeddings down to 2 and visualize them in Fig \ref{fig:embed}. Qualitatively, we see the model does fairly good job in separating the anomalous examples from the normal examples.

\begin{figure*}[!htp]
    \centering
    \subfloat[]{
        \includegraphics[scale=0.4, clip]{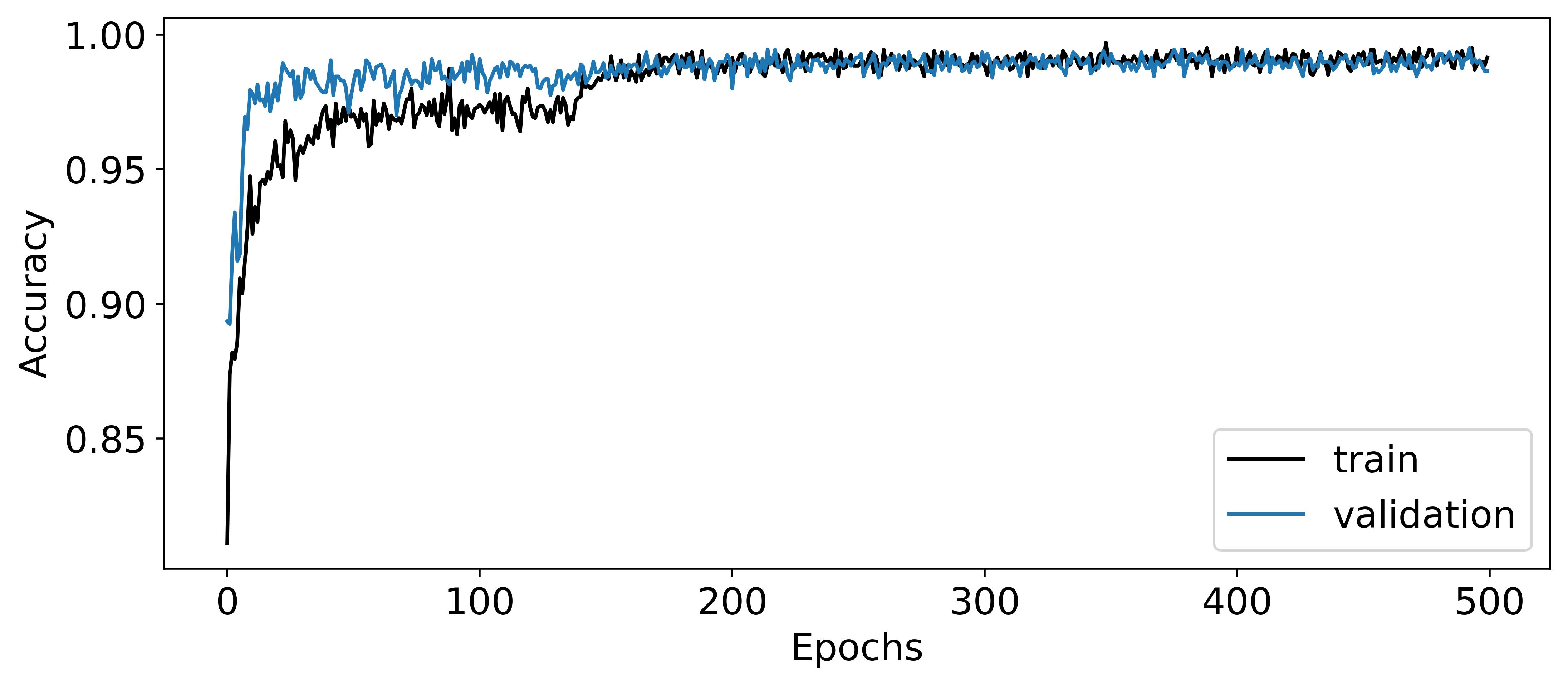}
    }
    \subfloat[]{
        \includegraphics[scale=0.4, clip]{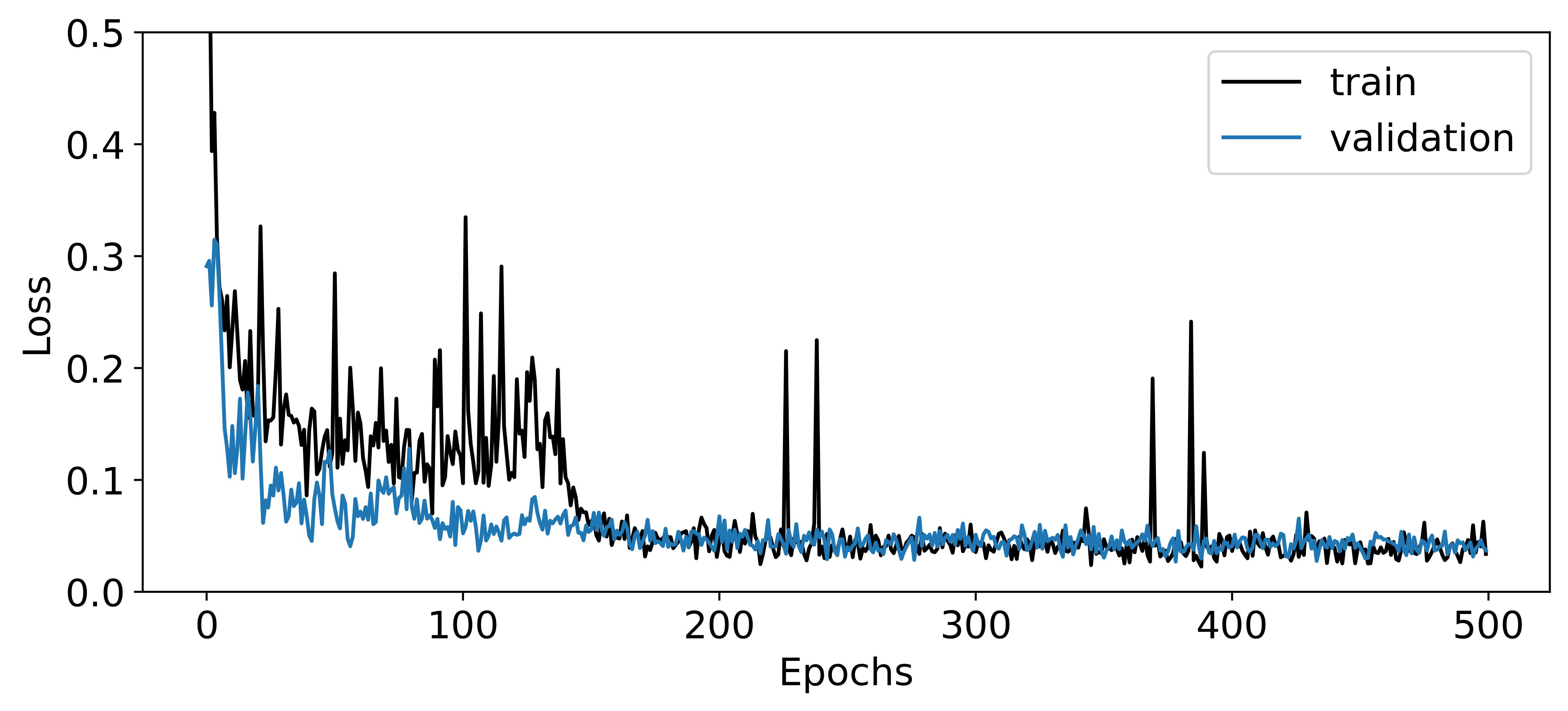}
    }
    \hspace{0mm}
    \subfloat[]{
        \includegraphics[scale=0.4, clip]{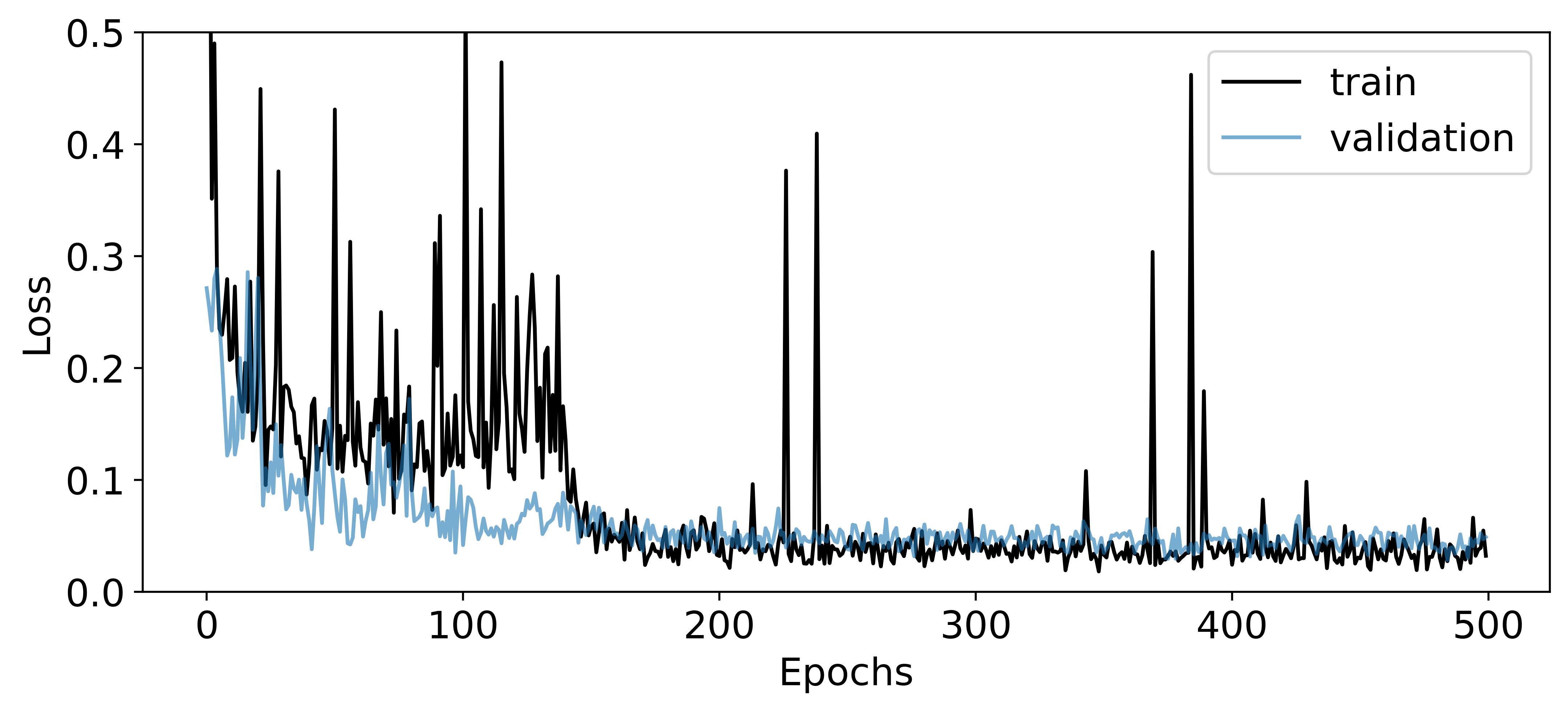}
    }
    \subfloat[]{
        \includegraphics[scale=0.4, clip]{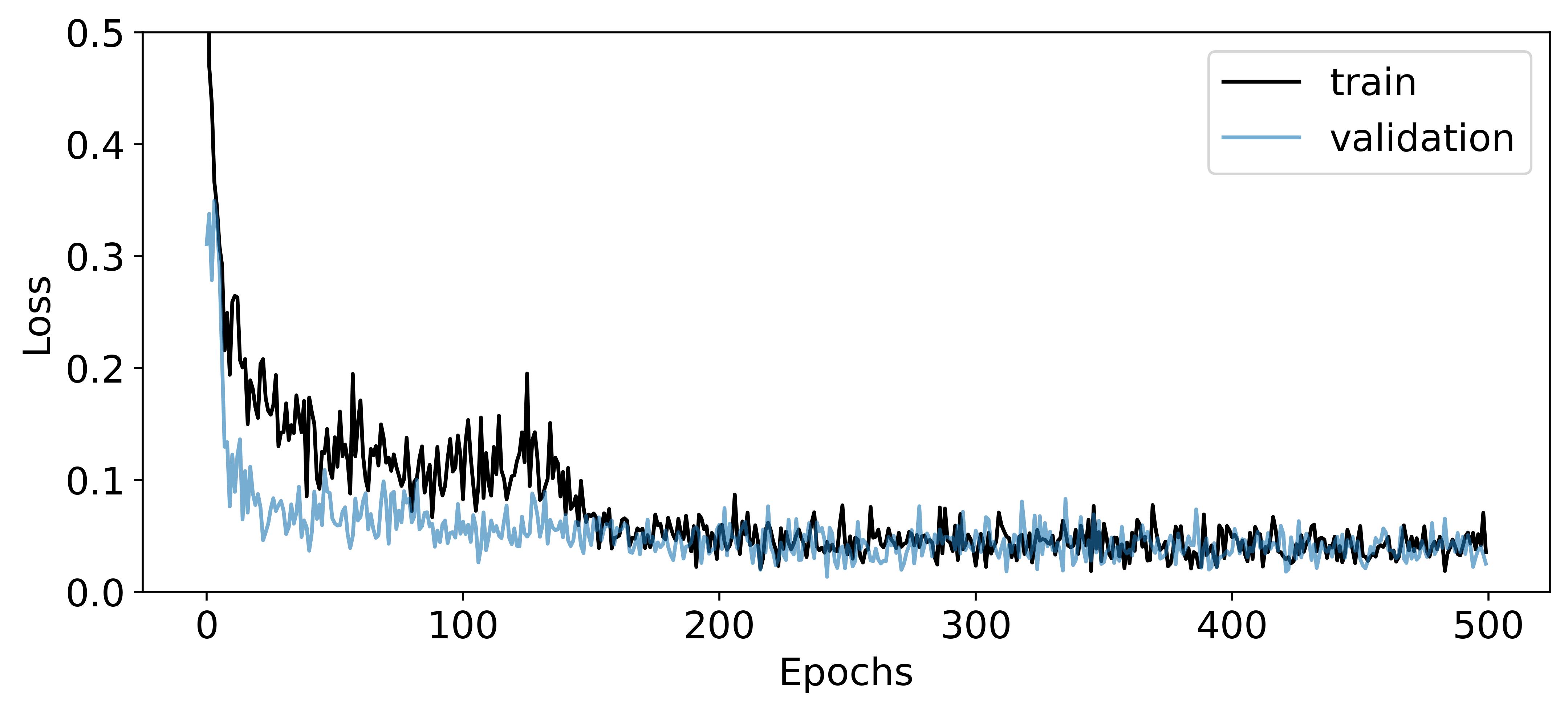}
    }

    \caption{Training regime of our proposed hybrid ProtoSiamese model. (a) Accuracy for different epochs (b) Overall loss reduction (c) The prototypical loss (d) Triplet loss}
    \label{fig:rslts}
\end{figure*}

\begin{table}[h]
    \centering
    \scriptsize
    \begin{tabular}{ll}\hline
         Model &  Accuracy (validation)\\\hline
         Baseline binary classifier & 98.5\% \\
         Baseline multi-class classifier & 95.3\% \\
         \textbf{Our hybrid protoSiamese} & \textbf{98.6\%} 
    \end{tabular}
    \caption{Comparing our proposed model with existing supervised classification tasks}
    \label{tab:cmprbsln}
\end{table}

\begin{figure}
    \centering
    \includegraphics[scale=0.5, clip]{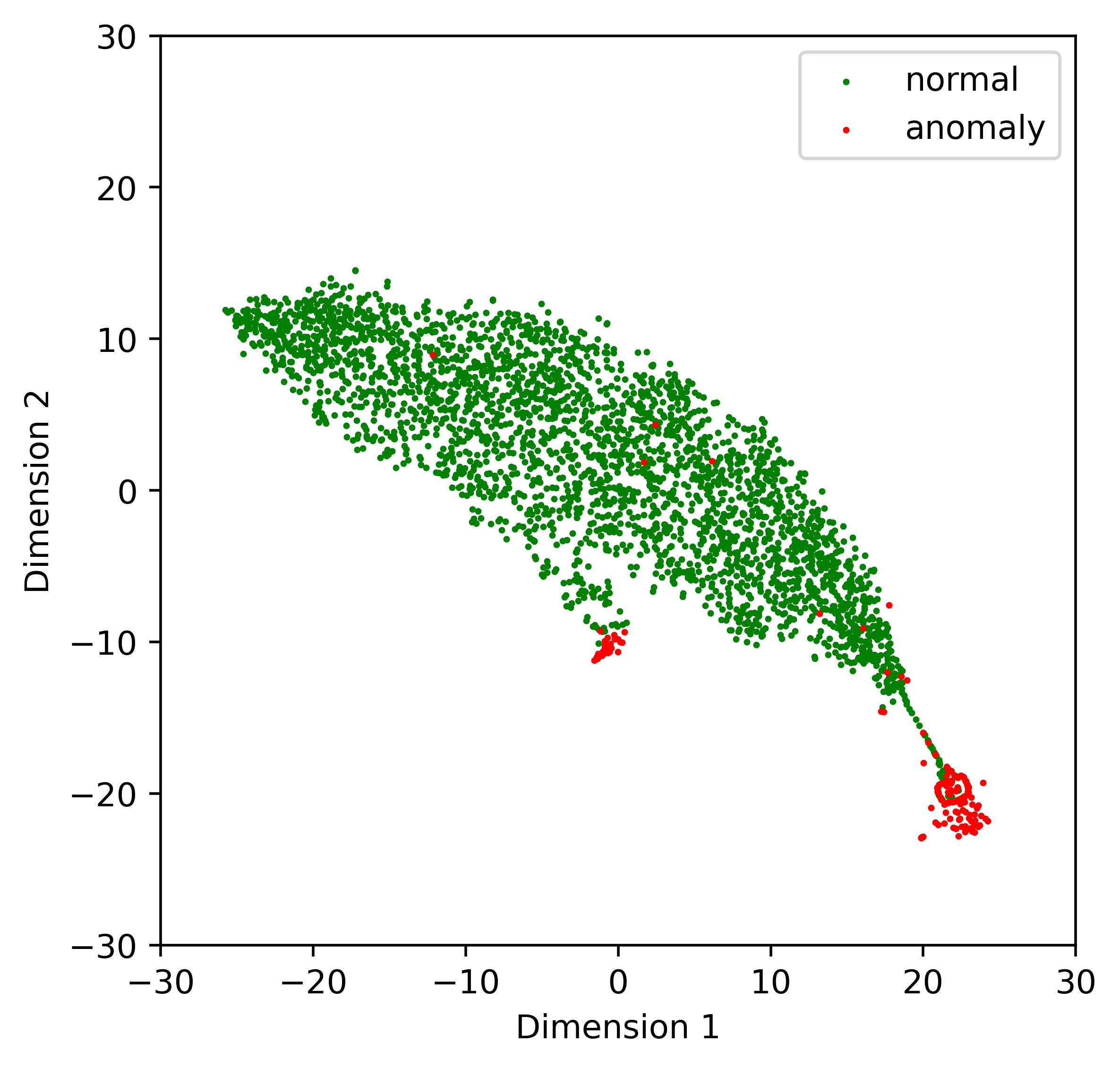}
    \caption{Visualization of the embeddings learned from the model. Compressed into 2-dimensions using TSNE}
    \label{fig:embed}
\end{figure}
\section{Conclusion}
Traditional machine learning or deep learning based approaches would not be a good choice for classification of low data or imbalance data problem for log anomaly detection. 
This paper proposed a meta learning based framework LogAnMeta, for log anomaly detection. This framework  combine two popular meta learning algorithms (Proto net, Siamese net) and train on very few samples, which extrudes the challenge of low data problem. Our model is able to learn new types of anomalies from only a small number of examples and the accuracy is also competitive. It outperforms the baseline binary and multi-class classifier. This method can be used to detect anomalies  from real time logs with very few anomalous samples with some minor modifications. Our model performance is affected by our choice to represent log sequences using a frequency count based approach which might lose some contextual semantic information. 

Future work includes Bidirectional Encoder Representations from Transformers (BERT) based sequence of log keys encoding to extract semantic information and use this encodings to train a hybrid few-shot classifier in an episodic manner.

\bibliographystyle{unsrt}
\bibliography{Reference}

\begin{thebibliography}{10}

\bibitem{zhu2019tools}
Jieming Zhu, Shilin He, Jinyang Liu, Pinjia He, Qi~Xie, Zibin Zheng, and
  Michael~R Lyu.
\newblock Tools and benchmarks for automated log parsing.
\newblock In {\em 2019 IEEE/ACM 41st International Conference on Software
  Engineering: Software Engineering in Practice (ICSE-SEIP)}, pages 121--130.
  IEEE, 2019.

\bibitem{he2017drain}
Pinjia He, Jieming Zhu, Zibin Zheng, and Michael~R Lyu.
\newblock Drain: An online log parsing approach with fixed depth tree.
\newblock In {\em 2017 IEEE international conference on web services (ICWS)},
  pages 33--40. IEEE, 2017.

\bibitem{he2016experience}
Shilin He, Jieming Zhu, Pinjia He, and Michael~R Lyu.
\newblock Experience report: System log analysis for anomaly detection.
\newblock In {\em 2016 IEEE 27th international symposium on software
  reliability engineering (ISSRE)}, pages 207--218. IEEE, 2016.

\bibitem{he2017towards}
Pinjia He, Jieming Zhu, Shilin He, Jian Li, and Michael~R Lyu.
\newblock Towards automated log parsing for large-scale log data analysis.
\newblock {\em IEEE Transactions on Dependable and Secure Computing},
  15(6):931--944, 2017.

\bibitem{xu2009detecting}
Wei Xu, Ling Huang, Armando Fox, David Patterson, and Michael~I Jordan.
\newblock Detecting large-scale system problems by mining console logs.
\newblock In {\em Proceedings of the ACM SIGOPS 22nd symposium on Operating
  systems principles}, pages 117--132, 2009.

\bibitem{du2017deeplog}
Min Du, Feifei Li, Guineng Zheng, and Vivek Srikumar.
\newblock Deeplog: Anomaly detection and diagnosis from system logs through
  deep learning.
\newblock In {\em Proceedings of the 2017 ACM SIGSAC conference on computer and
  communications security}, pages 1285--1298, 2017.

\bibitem{zhang2016automated}
Ke~Zhang, Jianwu Xu, Martin~Renqiang Min, Guofei Jiang, Konstantinos
  Pelechrinis, and Hui Zhang.
\newblock Automated it system failure prediction: A deep learning approach.
\newblock In {\em 2016 IEEE International Conference on Big Data (Big Data)},
  pages 1291--1300. IEEE, 2016.

\bibitem{zhang2019robust}
Xu~Zhang, Yong Xu, Qingwei Lin, Bo~Qiao, Hongyu Zhang, Yingnong Dang, Chunyu
  Xie, Xinsheng Yang, Qian Cheng, Ze~Li, et~al.
\newblock Robust log-based anomaly detection on unstable log data.
\newblock In {\em Proceedings of the 2019 27th ACM Joint Meeting on European
  Software Engineering Conference and Symposium on the Foundations of Software
  Engineering}, pages 807--817, 2019.

\bibitem{meng2019loganomaly}
Weibin Meng, Ying Liu, Yichen Zhu, Shenglin Zhang, Dan Pei, Yuqing Liu, Yihao
  Chen, Ruizhi Zhang, Shimin Tao, Pei Sun, et~al.
\newblock Loganomaly: Unsupervised detection of sequential and quantitative
  anomalies in unstructured logs.
\newblock In {\em IJCAI}, volume~19, pages 4739--4745, 2019.

\bibitem{koch2015siamese}
Gregory Koch, Richard Zemel, and Ruslan Salakhutdinov.
\newblock Siamese neural networks for one-shot image recognition.
\newblock In {\em ICML deep learning workshop}, volume~2. Lille, 2015.

\bibitem{snell2017prototypical}
Jake Snell, Kevin Swersky, and Richard~S Zemel.
\newblock Prototypical networks for few-shot learning.
\newblock {\em arXiv preprint arXiv:1703.05175}, 2017.

\end{thebibliography}

\end{document}